%% file: W19P07.tex
\documentclass[runningheads]{llncs}
\usepackage{graphicx}

\usepackage{comment}
\usepackage{amsmath,amssymb} 
\usepackage{color}
\usepackage[table]{xcolor}
\usepackage{subcaption}
\usepackage{multirow}
\newcommand{\norm}[1]{\left\lVert#1\right\rVert}
\newcommand{\dgr}{$^{\circ}$}

\begin{document}
\pagestyle{headings}
\mainmatter
\def\ECCVSubNumber{W19P07}  
\input{only_figures}

\title{Insights on Evaluation of Camera Re-localization Using Relative Pose Regression}

\titlerunning{Insights on Relative-Pose Regression}
%

\author{Amir Shalev\inst{1,2} \and
Omer Achrack\inst{2} \and
Brian Fulkerson \and 
Ben-Zion Bobrovsky \inst{1} 
}

%
\authorrunning{Shalev et al.}
%
\institute{Tel-Aviv-University  \email{Bobi@eng.tau.ac.il} \and Intel \email{\{Amir.Shalev,Omer.Achrack\}@intel.com} } 


\maketitle

\begin{abstract}
We consider the problem of relative pose regression in visual relocalization. Recently, several promising approaches have emerged in this area. We claim that even though they demonstrate on the same datasets using the same split to train and test, a faithful comparison between them was not available since on currently used evaluation metric, some approaches might perform favorably, while in reality performing worse. We reveal a tradeoff between accuracy and the 3D volume of the regressed subspace. We believe that unlike other relocalization approaches, in the case of relative pose regression, the regressed subspace 3D volume is less dependent on the scene and more affect by the method used to score the overlap, which determined how closely sampled viewpoints are. We propose three new metrics to remedy the issue mentioned above. The proposed metrics incorporate statistics about the regression subspace volume. We also propose a new pose regression network that serves as a new baseline for this task. We compare the performance of our trained model on Microsoft 7-Scenes and Cambridge Landmarks datasets both with the standard metrics and the newly proposed metrics and adjust the overlap score to reveal the tradeoff between the subspace and performance. The results show that the proposed metrics are more robust to different overlap threshold than the conventional approaches. Finally, we show that our network generalizes well, specifically, training on a single scene leads to little loss of performance on the other scenes.
\end{abstract}
\insertFigTradeoff
\section{Introduction}
Visual Simultaneous Localization and Mapping (V-SLAM) is a widely used method in vision-based applications, including mobile robots, virtual reality (VR), augmented reality (AR) and navigation (from domestic environments to rockets and satellites), as well as in many more applications in multifarious fields. Although GPS is a robust and readily available solution for localization, for many applications it is not affordable. Due to its fundamental importance as a core technology for a wide range of applications, visual localization has been extensively studied with major progress evidenced over the past few years \cite{murAcceptedTRO2015}, \cite{Engel2016DirectSO}, \cite{Engel2014LSDSLAMLD}, \cite{Brachmann2016DSACD}, \cite{Sattler2017EfficientE}.
However, accurate and robust localization is still a challenging problem, alongside the growing demands for constructing better maps. If an agent needs to locate itself when a map is given in addition to visual clues, the task is called re-localization \cite{kiddnaped_robot_problem}. A few examples of re-localization are: loop closure for autonomous navigation,  map loading for virtual and augmented reality, and the kidnap robot problem. Re-localization can be viewed as regressing the pose of the camera when prior knowledge is given as a map. In recent years, researchers have used the deep learning approach, and trained a deep neural network to regress the camera pose on a query (test) image, given a set of training images. One of the most effective known techniques is called \emph{relative-pose-regression with image retrieval}; it is computationally efficient and able to generalize to new  \textbf{unseen} scenes, as we will demonstrate herein. Other published works  \cite{Laskar2017CameraRB}, \cite{Balntas2018RelocNetCM}, \cite{Saha2018ImprovedVR}, \cite{Ding_2019_ICCV} present different implementations and techniques to improve re-localization performance. Detailed evaluations of these and other works were done independently  by \cite{valada2018deep}, \cite{Sattler2019UnderstandingTL}, \cite{shavit2019introduction}. 
However, to the best of our knowledge, our work is the first to take into account the elusive tradeoff that has been overlooked in the comparison criteria published to date. It transpires that choosing different parameters on the same model affects the accuracy of the model, when using the current metrics. In this work, we will demonstrate this tradeoff, both qualitatively and empirically.
This paper introduces new metrics that take into account the subspace size. To evaluate our new metrics we used a relative-pose-regression framework for camera re-localization as a baseline for future works. Our Siamese network consists of two equivalent backbones with shared weights. Each learns to encode geometric information from an image into a feature vector. In the test stage, only one of these networks is used to estimate the camera pose. This method is able to generalize well, as we will show \ref{sec:exp}; we train it on one scene and show that the learned model is able to work well on the other scenes without retraining. We also present a new technique for estimating the amount of correlated information, given a pair of images, based on their location and some prior knowledge on the environment.

To summarize, we offer the following contributions:
\begin{itemize}
  \item We raise a concern regarding erroneous conclusions drawn in published relative-pose-regression camera methods. We will show that due to neglect of the tradeoff we revealed, some approaches perform favorably in terms of current evaluation metrics, while, in reality, they perform badly.
  \item We propose to revise the metrics used for evaluating relative-pose-regression camera methods. We propose three metrics, which incorporate some statistics regarding the scene volume. Our new metrics depend on the 3D volume occupied and on how closely  viewpoints are sampled.
  \item We introduce a new way to estimate the amount of correlated information between pairs of images by 3D frustum-overlap. This approach makes the perception of the neglected tradeoff more intuitive.  
  \item We present a better way to use the overlap score, and we demonstrate the robustness and benefits of the relative-pose-regression method that produces competitive results on untrained environments.
\end{itemize}

The rest of the paper is organized as follows:
Sec.~\ref{sec:related_work} briefly reviews the related work in visual re-localization. The details of our approach, network structure, and overlap score function are detailed in Sec.~\ref{sec:method}.
Sec.~\ref{sec:comparison} is devoted to reviewing the comparison criteria. 
Evaluation results are provided in Sec.~\ref{sec:exp}.
In Sec.~\ref{sec:conclusion}, we summarize the results and conclusions, and offer some suggestions for future work.
Our source code will be publicly available soon.

\section{Related Work}
\label{sec:related_work}
\subsection{Re-localization approaches}
The task of camera re-localization has a long history of research  in various V-SLAM systems. Traditional approaches, as well as some leading algorithms \cite{Sattler2012ImprovingIL}, \cite{Brachmann2016DSACD}, \cite{GalvezTRO12}, are built on multi-view geometry theory. Other methods, like appearance-based similarity, the Hough Transform \cite{Tejani2014LatentClassHF}, and random-forest based methods \cite{Meng2017BacktrackingRF}, \cite{Shotton2013SceneCR}, \cite{GuzmnRivera2014MultioutputLF}, have been investigated and shown good results on some benchmarks. Recently, deep learning approaches have become a popular end-to-end solution for re-localization problems \cite{Shotton2013SceneCR}, \cite{gordo2016deep}, \cite{Radenovi2016CNNIR}, \cite{Larsson_2019_ICCV}, instead of only being a replacement for parts of the re-localization pipeline \cite{Brachmann2019ExpertSC}. The benefits of this approach are reduction of the inference time and of memory consumption, which is crucial for low computation and memory devices like drones, AR/VR and mobile devices. The major methods are now briefly summarized.
\subsubsection{Features-Based}
\label{sec:Related:features-based}
Methods are methods that use multi-view geometry features extractors and descriptors. This approach is the base for several leading solutions \cite{Brachmann2016DSACD}, \cite{Brachmann2017LearningLI}, \cite{Laskar2016RobustLC}, \cite{Sattler2012ImprovingIL}, \cite{Schnberger2017SemanticVL}. However, this family of methods has a major drawback: they are limited to a small working area since the computational costs grow significantly with working area size, even after optimization, such as is found in the case presented in \cite{Engel2014LSDSLAMLD} and others.
\subsubsection{Fiducial Markers}
\label{sec:Approaches:Markers}
One of the most popular approaches is based on the use of binary square fiducial markers. The main benefit of these markers is that a single marker provides enough correspondences to obtain the camera pose. Every marker is associated with a coordinate system, and poses are given relative to that coordinate system's origin \cite{Wang2016AprilTag2E}, \cite{Pfrommer2019TagSLAMRS}, \cite{Fiala2005ARTagAF}.

\subsubsection{Absolute Pose-Regression} 
\label{APR} 
The method outlined in \cite{PoseNet2015} suggested learning the re-localization pipeline in an end-to-end supervised manner. The idea of regressing the absolute pose by using machine learning offered several appealing advantages compared to traditional feature-based methods. Deep learning based on absolute pose-regression does not require the design of hand-crafted features. The trained model has a low memory footprint and constant runtime. However, any solution for absolute pose-regression suffers from over-fitting on the training data and will not generalize well - which is a desired property in practical cases.
Recently, many learning-based algorithms have been developed \cite{Cavallari2017OntheFlyAO}, \cite{Svrm2017CityScaleLF}, \cite{Massiceti2016RandomFV}, \cite{Meng2017ExploitingPA}.

\subsubsection{Relative-Pose-Regression with Image Retrieval}
The key idea of relative-pose-regression is that once an anchor image has been determined, one can directly regress the relative camera pose between the anchor and test images, and thus obtain the absolute pose. The relative-pose-regression approach requires a definition of similarity. This definition has been proposed and studied recently in several works. The method enjoys many of the advantages of the previously described methods, \eg attractive computational costs stemming from its maintaining a relatively small database, decoupling of the pose-regression process from the coordinate frame of the scene. In addition, it does not require scene-specific training, as we will demonstrate. \cite{Laskar2017CameraRB}, \cite{Meng2017BacktrackingRF}, \cite{Sattler2017EfficientE}, \cite{Taira2018InLocIV}, \cite{Ding_2019_ICCV}, \cite{Melekhov2017RelativeCP}, \cite{Saha2018ImprovedVR}.

\subsubsection{Combination of Methods} 
The authors in \cite{valada2018deep} presented a multi-task training approach, leveraging relative-pose information during the training, and demonstrated impressive results.
Alternatively, a variety of combinations are outlined in \cite{Brahmbhatt2017GeometryAwareLO}, where, for sequence localization they combine absolute pose-regression from the current frame to relative pose from the previous frame. In addition, a combination of the new approaches with traditional algorithms has improved the results, such as where the combination was done using visual odometry algorithms, which take as input IMU and GPS sensors. In the test stage, predictions were further refined with pose graph optimization.

\subsection{Images Intersection Score}
In relative-pose-regression a metric is used to estimate the correlated information between pairs of images, i.e. to determine whether a given pair of images contains enough overlapped information such that the algorithm can retrieve the relative pose between them. A few metrics have evolved during the last few years: the authors in
\cite{Laskar2017CameraRB} measured the overlap as the percentage of pixels projected onto a candidate image plane using the respective ground truth pose and depth maps; the method in
\cite{Balntas2018RelocNetCM} projects depth from one pose, and un-projects it from another, and then counts how many points are on the image plane; the authors in
\cite{Ding_2019_ICCV} suggested using ORB similarity for outdoor scenes with 2D image data-sets. Both overlap calculation methods use the geometric as well as the visual content of the images. In this work, we propose using pose information from the ground-truth, and do not assume depth information is available.
Although each method has its advantages, we think that none provides adequate intuitive understanding of the sub-space that is spread by the span of all the relative pose. To this end we now introduce a new, and more intuitive, image intersection score we call frustum-overlap.

\section{Methodology}
\label{sec:method}
The main novelty of this work is to revise the problem of re-localization assessment expressed with standard comparison criteria, and to propose a new comparison protocol. We therefore used the popular models and methods that are known to work on the problem, such as camera re-localization. 
\subsection{Frustum-Overlap - An Intuitive Images Intersection Score}
\label{sec:ImagesPairIntersection:FrustumOverlap}
\begin{figure}[h]
\begin{center}
    \begin{subfigure}[b]{0.28\textwidth}
       \includegraphics[width=0.95\linewidth]{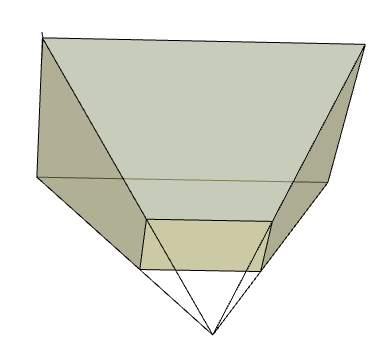}
       \caption{Frustum Type a}
       \label{fig:Frustum_Type_a}
    \end{subfigure}
    \begin{subfigure}[b]{0.28\textwidth}
       \includegraphics[width=0.95\linewidth]{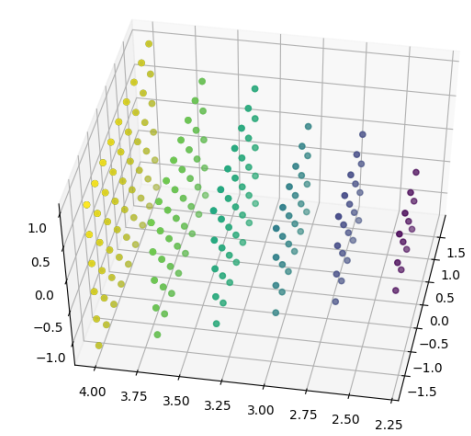}
       \caption{Frustum Type b}
       \label{fig:Frustum_Type_b}
    \end{subfigure} 
    \begin{subfigure}[b]{0.20\textwidth}
        \includegraphics[width=0.99\linewidth]{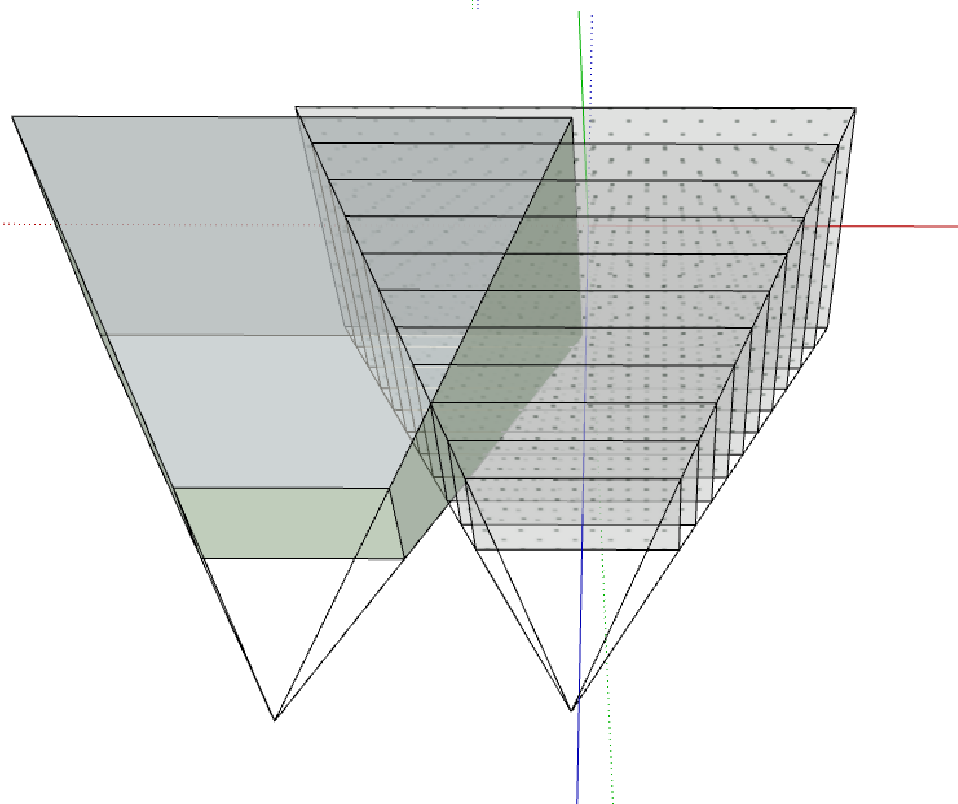}
        \caption{}
        \label{fig:overlap_definition:c}
    \end{subfigure} 
    \begin{subfigure}[b]{0.20\textwidth}
        \includegraphics[width=0.99\linewidth]{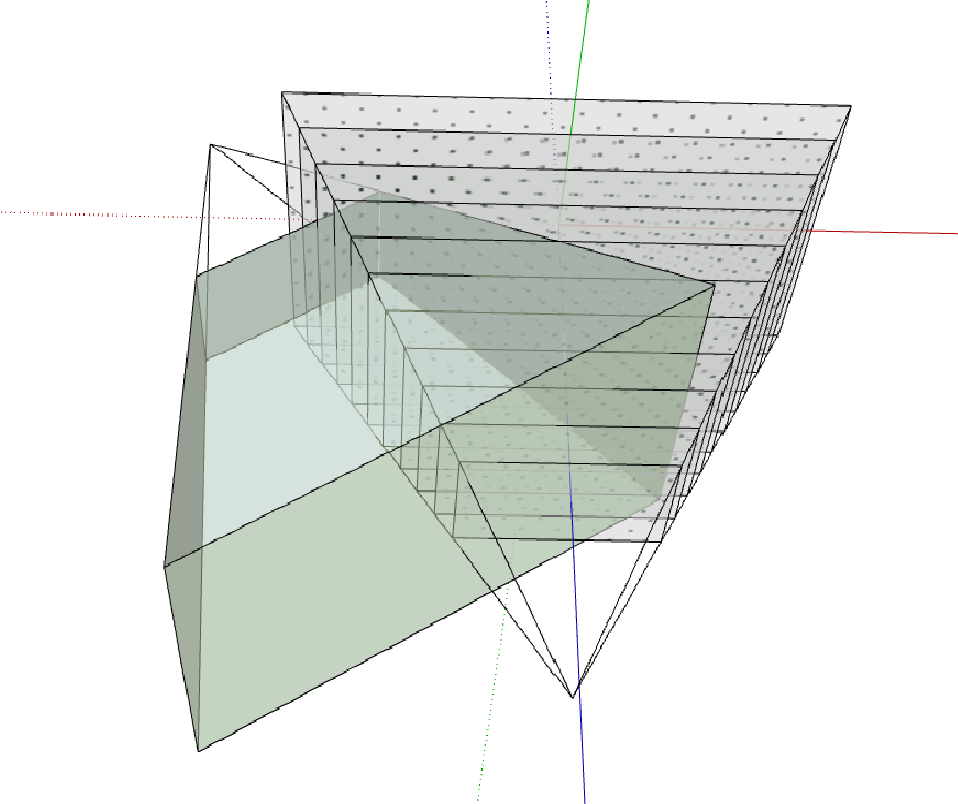}
        \caption{}
        \label{fig:overlap_definition:d}
    \end{subfigure}
\caption{Frustum types}
\label{fig:Frustum_Types}
\end{center}
\end{figure}
We introduce a novel definition for image overlap that does not rely on the content of the images. We borrowed the concept of the camera view frustum from computer graphics. A frustum is a truncated pyramid that starts at the focal plane of the camera, and extends up to the maximum viewable distance. We use two different ways to define a frustum: 
\begin{itemize}
    \item \textbf {Type a} is presented in Fig.~\ref{fig:Frustum_Type_a} - A set of 6 3D planes, 2 parallel planes and 4 planes delimiting them to a pyramid.
    \item \textbf{Type b} is presented in Fig.~\ref{fig:Frustum_Type_b} - A discrete set of 3D points on a grid, constructing a frustum shape.
\end{itemize}
To get an overlap score given a pair of two poses, we need to place a frustum of {\it Type a} in the first pose and a frustum of {\it Type b} in the other pose, and then extract the percentage of 3D points of the second frustum that are inside the first frustum. This is illustrated in Fig.~\ref{fig:overlap_definition:c}. In Equation ~\ref{overlap_count} we describe the counting formula we used, to give some informal intuition behind the calculation.
\begin{equation}
    \begin{split}
         &overlap\_score =  \sum_{p_i \in frustum\_a} \frac{a_i}{N_b} \\
         &a_i=\begin{cases}
            1, &  \text{if $p_i$ is in frustum\_b}\\
            0, & \text{otherwise}\\
          \end{cases} \\
    \end{split}
    \label{overlap_count}
\end{equation}
\begin{center}
     Our overlap scoring method. $N_b$ is the total number of points in frustum b.
\end{center}
This approach has a major drawback - it is possible to get a high overlap score for two images without any common visual content (see Fig.~\ref{fig:overlap_definition:d} for an illustration.) We overcome this drawback by limiting the relative rotation between two captures.
It is important to emphasize that, in contrast to other methods, our method scores overlap in 3D space, and therefore has a tight relationship with the span subspace. A formal definition of the calculation of the score is given in the supplementary material.    
\begin{figure*}
    \begin{center}
       \includegraphics[width=0.50\linewidth]{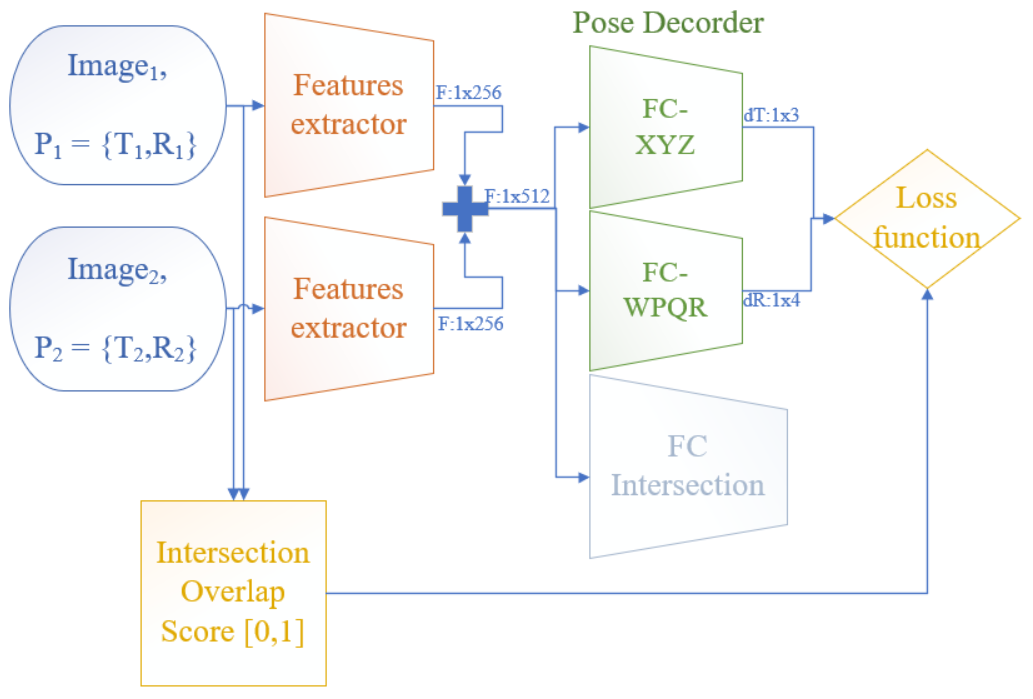}
    \end{center}
    \captionsetup{width=.9\textwidth}
    \caption{Diagram of the proposed training architecture: each Siamese branch uses a shared features extractor to get a representation of the image. The pose decoder predicts rotation, represented by a quaternion, and translation, using ${x,y,z}$ coordinates.}
    \label{fig:architecture}
\end{figure*}
\subsection{Training Procedure}
Our training procedure is done in a straightforward way. Given the pre-defined pairs of images induced by the selection of the overlap computing method and overlap threshold, as well as their corresponding relative pose as a label, we leverage Siamese architecture to get the estimated relative pose for the network. A schematic of our training procedure is given in Figure \ref{fig:architecture}.
\subsection{Loss Function}
Loss definition for supervised pose-regression tasks is challenging because it involves learning two distinct quantities - rotation and translation. Each has its own different units and different scales. The authors in \cite{Kendall2017GeometricLF} found that outdoor and indoor scenes are characterized by different weights, which motivates to dynamically learn the weights of the loss for each task instead of determining them in a hard-coded manner. We follow this notion and use learned $\alpha$ and $\beta$ to balance the losses and combine them, as described formally in \cite{Kendall2017GeometricLF}:
\begin{subequations}
\begin{align}
&L_{q}(\hat{q_{rel}},q_{rel}) = \norm{q_{rel} - \frac{\hat{q_{rel}}}{\norm{{\hat{q_{rel}}}}_2}}_2 \\
&L_{t}(\hat{t_{rel}},t_{rel}) = \norm{t_{rel} - {\hat{t_{rel}}}}_2\\
&L(\hat{p_{rel}},p_{rel}) =  \alpha^2 + \beta^2 + e^{-\alpha^2 } L_t(\hat{t_{rel}},t_{rel}) +  e^{-\beta^2} L_q(\hat{q_{rel}},q_{rel}) 
\end{align}
\end{subequations}
where $t_{rel}$ and $q_{rel}$ are the relative translation and rotation of the $\hat{}$ superscript that denotes predicted values.
This form of loss function makes it possible for us to learn the weighting between the translation and rotation objectives and to find optimal weighting for a specific data-set.

\section{Comparison Criteria}
    \label{sec:comparison}
    \subsection{Standard Comparison Criteria}
    \label{sec:comparison:Conventional}
    To evaluate the performance of the proposed algorithm, we require a set of images and their corresponding ground truth poses. We estimate the image poses predicted by the algorithm and calculate the pose error for translation and rotation. The popular localization metrics are:   
    \begin{subequations}
    \begin{align}
    &t_{err}(\hat{t}_i,t_i)= \Vert{t_i - \hat{t_i}}\Vert_2 \\
    &q_{err}(\hat{q}_i,q_i)= \frac{180}{\pi}2\cos^{-1}(<q_i,\hat{q_i}>)
    \end{align}
    \end{subequations}
    After obtaining a set of translation and rotation errors, a statistical measure to convert them to a single scalar score is used. Common choices \cite{Brahmbhatt2017GeometryAwareLO}, \cite{PoseNet2015}, \cite{Balntas2018RelocNetCM} are the mean and the median:
    \begin{subequations}
    \begin{align}
    &T_{err} = mean/median(\{t_{err}(\hat{t}_i,t_i)\}_{i \in N})\\
    &Q_{err} = mean/median(\{q_{err}(\hat{q}_i,q_i)\}_{i \in N})
    \end{align}
    \label{eqn:comp_criteria}
    \end{subequations}
    
    
    \subsection{The limitations of standard comparison criteria}
    
    \label{sec:comparison:issues}
    In absolute-pose-regression, when one calculates the error using Equation \ref{eqn:comp_criteria} the scale (regression subspace volume) is implicitly involved in the computation. The error is limited by the scene dimensions, and can be manipulated by normalizing the scene. Moreover, the train and test sets are exactly the same, because there is no \textbf{pre-defined overlap based selection}. In relative-pose-regression, this is not the case; the selection of the overlap method and threshold implicitly determines the complexity of the problem, errors are limited by this selection and scene normalization will not solve it. Changing either the overlap calculation method or the threshold might significantly change the train and test data. An illustration of this key insight is depicted in Figure \ref{fig:tradeoff}. Qualitatively, if one chooses 'high' overlap, the regression subspace volume is smaller than if one choses 'low' overlap. This, in turn, will lead at first sight to the conclusion that the algorithm that trained and tested on 'high' overlap is better, but we argue that this is not necessarily so. We further argue that there is a strong correlation between the spanned 3D volume induced by the overlap selection and the performance of the selected algorithm, as we will show in section \ref{sec:exp}. We demonstrate this claim visually in Figure \ref{fig:Naive-Estimation-RMSE}
    
    \insertFigNaiveEstimationRMSE
    \insertTableBeatThemAllRPR
    \insertTableBeatThemAllAPR
    
    \subsection{Proposed Metrics}
    \label{sec:metric:agnostic}
        As a direct sequitur to the previous sections, the results obtained using a naive implementation of Equation \ref{eqn:comp_criteria} is not enough to make a fair comparisons. As mentioned earlier, we claim that in regression methods comparison with Equation \ref{eqn:comp_criteria} is reliable only if their volume is similar. However, in the case where we do not know the subspace volume we need other information in order to get a reliable comparison. Hence, our first suggestion is to use a sub-space agnostic metric. As a minimal requirement for fair comparison, one should include the subspace volume or a statistical representation of it.
        \subsubsection{Statistical Criteria Inspired from Financial Analysis}
\begin{figure}[t]
            \begin{center}
                \begin{subfigure}[b]{0.32\textwidth}
                    \includegraphics[width=0.99\linewidth]{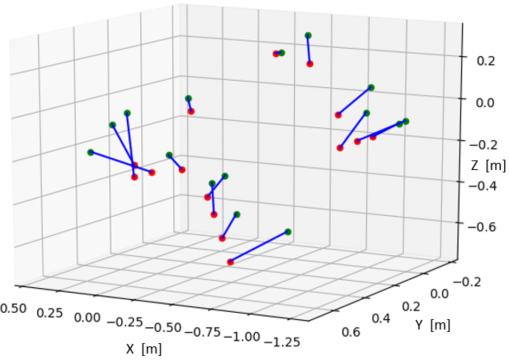}
                    \captionsetup{width=0.8\textwidth}
                    \caption{Translation error vector}
                    \label{fig:visualize_metric:median}
                    \end{subfigure} 
                \begin{subfigure}[b]{0.32\textwidth}
                    \includegraphics[width=0.99\linewidth]{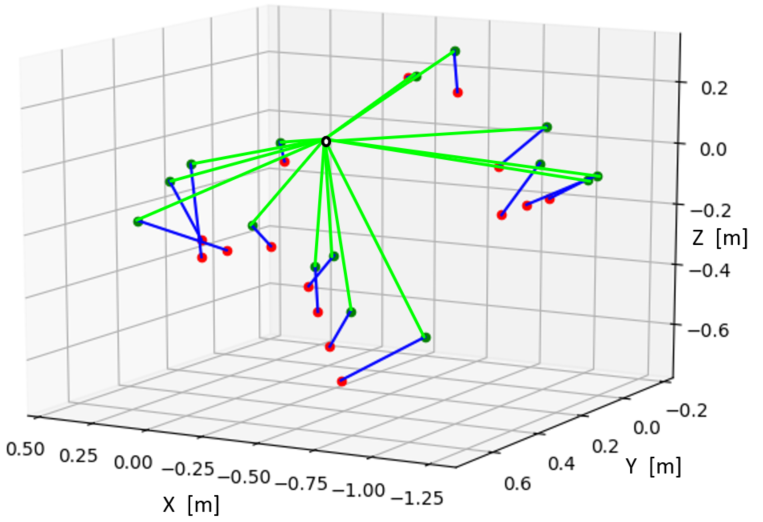}
                    \caption{Mean Absolute Percentage Error (MAPE)}
                    \label{fig:visualize_metric:MAPE}
                \end{subfigure} 
                \begin{subfigure}[b]{0.32\textwidth}
                    \includegraphics[width=0.99\linewidth]{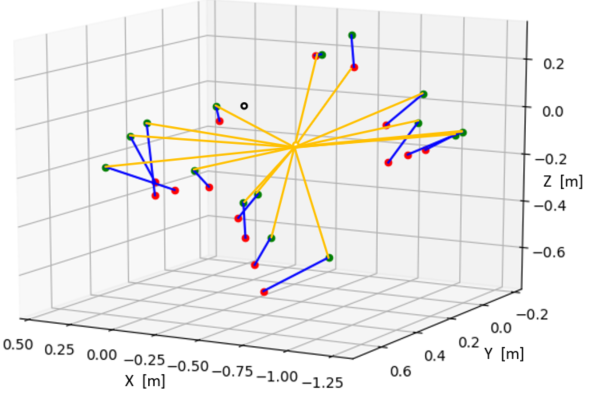}
                    \caption{Mean Absolute Scale Error (MASE)}
                    \label{fig:visualize_metric:MASE}
                \end{subfigure} 
                \end{center}
                \captionsetup{width=.9\textwidth}
                \caption{An Illustration of the proposed metrics. Ground truth points are in green, estimation points in red, the  blue line connected them is the translation error. Green and yellow lines represent the normalization factor of the error in MAPE and MASE, respectively. These types of normalization are necessary when subspace volume is taken into account.} 
                \label{fig:visualize_metric} 
            \end{figure}

            \begin{itemize} 
            \item {\bf The mean absolute percentage error (MAPE)}, defined as:
            \begin{equation}
                T_{MAPE} = \frac{1}{N}\sum_{i \in N} (\frac{|t_i - \hat{t_i}|_1}{|t_i|_1})
            \end{equation} 
            
            The intuition behind this metric is: penalty decreases proportionally with the distance from the origin. Hence, it implies a larger penalty on errors in 'smaller' scenes \cite{2016_MAPE_for_regression}.\newline
            
            \item {\bf The mean absolute scaled error (MASE)} is another relative measure of the error. It is defined as the mean absolute error of the model divided by the mean absolute error of a naive random-walk-without-drift model (\ie the mean absolute value of the first difference of the series) \cite{MASE_2006}. Thus, it measures the relative error compared to a naive model:\footnote{By naive model we mean that the model always returns the mean value of the training data.}
            \begin{equation}
                T_{MASE\_error} = \frac{ T_{MAE}}{ T_{Naive}} = \frac{\sum_{i \in N} (|t_i - \hat{t_i}|)}{\sum_{i \in N} {|t_i - mean(\{|t|\}_{i \in N}|}}
            \end{equation}
            \item {\bf The mean absolute percentage scaled error (MAPSE)} is a combination of the two measures mentioned above. The benefits of using this measure are better normalization and scale-less errors:
            \begin{equation}
                T_{MAPSE\_error} = \frac{ T_{MAPE}}{ T_{Naive}} = \frac{\sum_{i \in N} (\frac{|t_i - \hat{t_i}|}{|t_i|})}{\sum_{i \in N} {|t_i - mean(\{|t|\}_{i \in N}|}}
            \end{equation}
            \end{itemize}
            We visually illustrate these measures in Figure \ref{fig:visualize_metric}
            
            The claims are valid for both rotation and translation. Subspace volume is not intuitive in quaternions representation, but if we convert to Euler angles we can use the same formulas. The subspace of absolute orientation in most scenes is the full sphere, but for the rotation (relative orientation) it is also much smaller. The "MAPE" formula is given by:
            \begin{equation}
                R_{MAPE}(Euler Angles) = \frac{1}{N}\sum_{i \in N} (\frac{|r_i - \hat{r_i}|}{|r_i|})
            \end{equation}
            
            \insertFigAUC
        
        \subsubsection{Area under curve for all overlap ranges}
        The accuracy of a model depends on how well it estimates the relative pose given a test pair of images. As mentioned earlier, the selection of the overlap method and threshold significantly determines the complexity of the problem. In order to achieve a non-overlap dependence on assessment, one way is to compute the area under the curve (AUC) across the overlaps.  The model is better as the area decreases. Quantification of the difference between the two models could be obtained from the difference between their AUCs. This is illustrated in Fig \ref{fig:AUC}. 
        
        \subsubsection{Minimal requirement for reliable comparison}
        As a relaxation of the previous method, it might be enough to consider the sub-space diameter\footnote{In \cite{Laskar2017CameraRB} a hard-coded metric was used. This provided correct predictions by limiting the translation and rotation errors up to a pre-defined threshold. This metric can be used in addition to considering the sub-space diameter}. Intuitively, the subspace diameter quantifies the spanned 3D space induced by the relative pose targets (labels).
        The sub-space diameter can be defined as:
        \begin{equation}
            subspace\_diameter=mean(\{|t|\}_{i \in N}) + 2 {std(\{|t|\}_{i \in N})}
        \end{equation}
        
        \insertTableOverlapDiameterSmall
        \insertFigMAPEAcrossOverlaps

\section{Experiments and Results}
\label{sec:exp}

\subsubsection{Comparison to other methods}: To demonstrate our key insights we conducted several experiments. We chose the 7-Scenes data-set \cite{Shotton2013SceneCR} and trained our architecture (see section \ref{sec:method}). As its name implies, this data-set consists of 7 different scenes taken using a Microsoft Kinnect camera. For each of these scenes, we compare our trained network using an appropriate overlap threshold to achieve satisfactory results. These experiments are summarized in Table \ref{table:beat_them_all}. It is worth noting that our comparison is not complete. The train and test data are \textbf{not} the same, due to pairs selection induced by the overlap method. At first sight, our method yields the best results compared to other methods. However, one should note that we chose the overlap accordingly in order to achieve better results using the common criteria metrics. This complies with what was argued in \ref{sec:comparison:issues}.  

\subsubsection{Comparison to relative-pose methods}:
We compare our method using the standard median to other relative-pose methods. Note that these comparisons are made without relating to the sub-space diameter, and are summarized in Table \ref{table:beat_them_all:RPR}.
\subsubsection{Correlation between sub-space diameter and accuracy}:
To analyze the correlation with sub-space diameter as defined in Section \ref{sec:comparison}, we measure the sub-space diameter for each scene over a range of different overlaps. We introduce here another data-set we used, called Cambridge \cite{PoseNet2015}. Results are summarized in Table \ref{table:overlap_diameter}. To prove empirically our claim  that there is a strong correlation between overlap and accuracy, we trained the Chess scene from the 7-scenes data-set with an overlap of 0.6 and tested it under various overlap ranges. Results are shown in Figure \ref{fig:diameter_accuracy_corr}.
\subsubsection{Volume agnostic methods}:
We offer sub-space diameter agnostic metrics. By agnostic, we mean that the proposed measure should be \textit{stable} under different overlap and threshold selection methods. We used our trained model on the Chess scene from the 7-scenes data-set. We evaluated the methods proposed in section \ref{sec:metric:agnostic} and showed that the metrics result is stable around the same values, even when we modified the overlap threshold. These results are summarized in Figure \ref{fig:agnosic_vloume_methods_3}. We also changed the common definition of MAPE and MASE, and used $L2$ as a norm to examine the robustness of our method to norm selection. Results are summarized in Figure  \ref{fig:agnosic_vloume_methods_4}.
\insertFigAgnosicVloumeMethodsIII
\insertFigAgnosicVloumeMethodsIV
\insertTableCompareSubspaceDiameter
\subsubsection{Comparison given sub-space diameter}:
As we mentioned earlier, a minimal requirement for fair comparison is referred to as the subspace diameter, as reported in \cite{Saha2018ImprovedVR}. Hence, we compare our method to theirs, as summarized in Table \ref{tab:compare_sub_space_diameter}.

\section{Conclusions}
\label{sec:conclusion}
In this paper, we introduce a novel comparison criterion for relative-pose-regression. We demonstrate the insufficiency of the current standard metrics in terms of their capability to assess relative-pose-regression algorithms. We also show that our proposed metrics overcome many of the drawbacks of existing metrics, and believe that the proposed metrics should be adopted in the field. We believe that relative-pose-regression is crucial for solving the problem of camera re-localization. Generalization (training on one scene and achieving good results on another) is a key to the widely-adopted full solution of the problem; however, we leave its extensive investigation for future work.    

 In our opinion, due to the lack of generalization capabilities of the absolute pose-regression approach, relative-pose-regression is now the most promising research direction in data-driven visual re-localization.
\par\vfill\par

\clearpage
%
%
\bibliographystyle{splncs04}
\bibliography{egbib}
\end{document}

%% file: only_figures.tex
\newcommand{\insertFigSiameseNetworks}{
\begin{figure*}[hbt!]
    \begin{center}
           \includegraphics[width=0.7\linewidth]{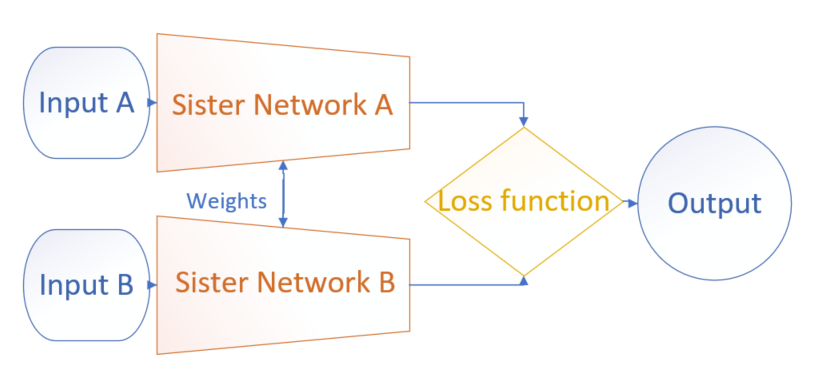}
    \end{center}
       \caption{Example of siamese networks. The two copies of the sister-network  share the same set of parameters.}
\end{figure*}
}
\newcommand{\insertFigTradeoff}{
\begin{figure}[hbt!]
    \begin{center}
        \begin{subfigure}{0.45\textwidth}
            \includegraphics[width=0.9\linewidth]{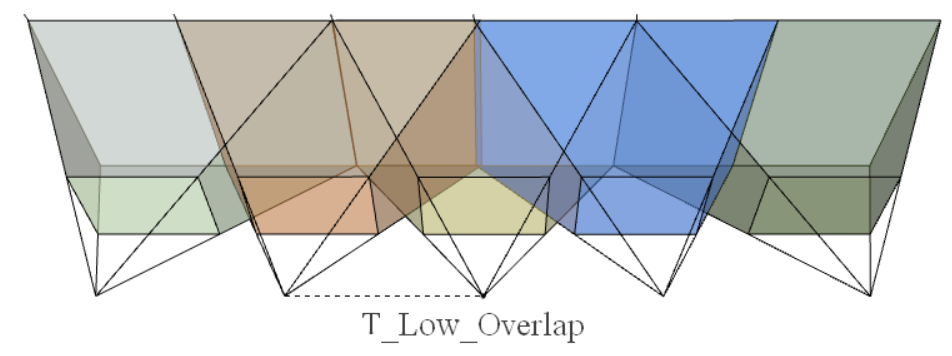}
            \caption{Low overlap spacing}
        \end{subfigure}
        \begin{subfigure}{0.45\textwidth}
            \includegraphics[width=0.87\linewidth]{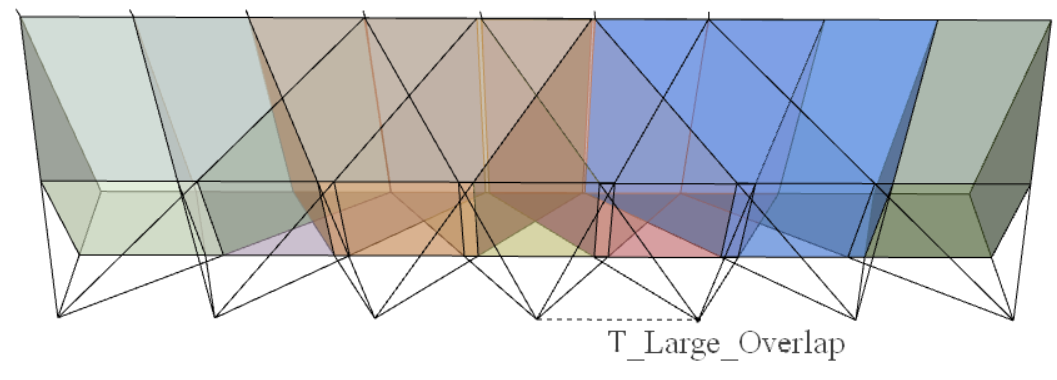}
            \caption{High overlap spacing}
        \end{subfigure}
    \end{center}
    \captionsetup{width=.95\textwidth}
    \caption{The figure illustrates a 3D space occupied by low (a) and high (b) Overlap spacing. Low overlap creates sparse sampling, which leads to a 'hard' relative-pose-regression problem. High overlap creates dense sampling, which leads to an 'easy' relative-pose-regression problem. The complexity of a regression problem is affected by the regression subspace volume. In terms of absolute-pose-regression, both are equivalent since we use the same space.}
    \label{fig:tradeoff}
\end{figure}
}
\newcommand{\insertFigFrustumTypes}{
\begin{figure}[hbt!]
    \begin{center}
        \begin{subfigure}{0.45\textwidth}
           \includegraphics[width=0.80\linewidth]{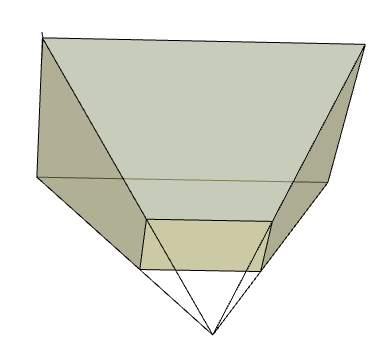}
           \caption{Frustum type {\it a}}
           \label{fig:Frustum_Type_a}
        \end{subfigure}
        \begin{subfigure}{0.45\textwidth}
           \includegraphics[width=0.80\linewidth]{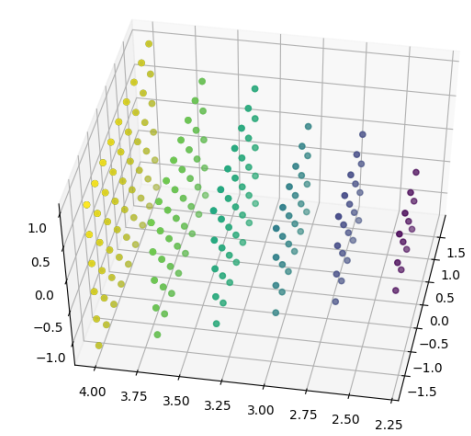}
           \caption{Frustum type {\it b}}
           \label{fig:Frustum_Type_b}
        \end{subfigure} 
    \end{center}
    \caption{Frustum types}
    \label{fig:Frustum_Types}
\end{figure}
}
\newcommand{\insertFigFrustumTypesB}{
\begin{figure}[h]
    \begin{center}
        \begin{subfigure}{0.45\textwidth}
           \includegraphics[width=0.99\linewidth]{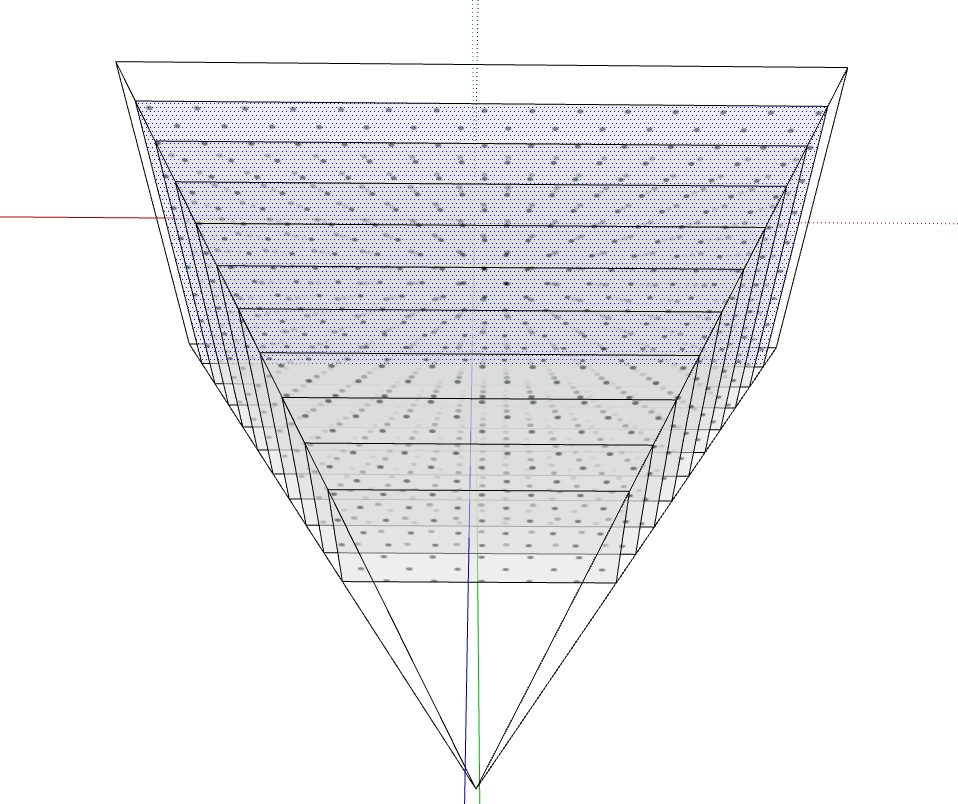}
           \caption{}
           \label{fig:overlap_definition:a}
        \end{subfigure}
        \begin{subfigure}{0.45\textwidth}
           \includegraphics[width=0.99\linewidth]{images/overlap_duble.png}
           \caption{}
           \label{fig:overlap_definition:b}
        \end{subfigure} 
        \begin{subfigure}{0.45\textwidth}
           \includegraphics[width=0.99\linewidth]{images/overlap_duble_150degree.png}
           \caption{}
           \label{fig:overlap_definition:c}
        \end{subfigure}
    \end{center}
    \captionsetup{width=.95\textwidth}
    \caption{{\bf(a)} type {\it b} frustum. {\bf(b)} Overlap calculation example. {\bf(c)} Demonstration of the importance of the orientation threshold.}
    \label{fig:overlap_definition}
\end{figure}
}
\newcommand{\insertFigFrustumTypeAVertex}{
\begin{figure*}[hbt!]
    \begin{center}
           \includegraphics[width=0.7\linewidth]{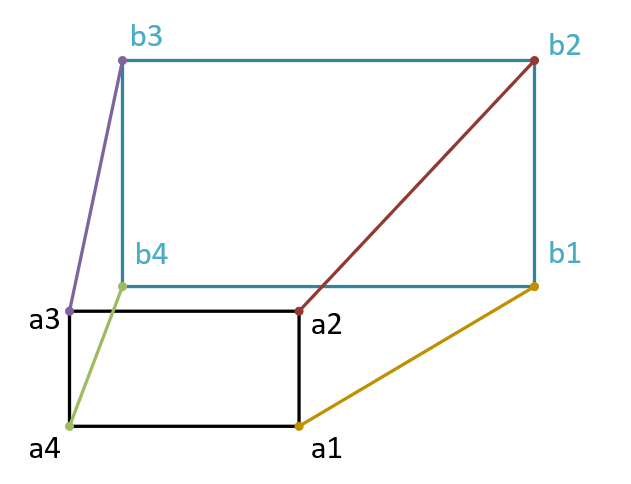}
    \end{center}
    \captionsetup{width=.95\textwidth}
    \caption{8 $3D$ points define frustum type a}
    \label{fig:Frustum_Type_A_vertex}
\end{figure*}
}
\newcommand{\insertTableSevenSceneDatasetSummery}{
\begin{table*}[hbt!]
\centering
 \begin{tabular}{|| l||c c c c||} 
  \hline\hline
    Scene & Train set & Test set & Spatial Extent [m] & Volume $[m^3]$ \\
  \hline\hline 
  Chess       & 4000 & 2000 & 3x2x1     & 6 \\
  Fire        & 2000 & 2000 & 2.5x0.5x1 & 1.25 \\
  Heads       & 1000 & 1000 & 2x0.5x1   & 1 \\
  Office      & 6000 & 4000 & 2.5x2x1.5 & 7.5 \\
  Pumpkin     & 4000 & 2000 & 2.5x2x1   & 5 \\
  Red Kitchen & 7000 & 5000 & 4x3x1     & 18 \\
  Stairs      & 2000 & 1000 & 2.5x2x1.5 & 7.5 \\
 \hline\hline
 \end{tabular}
 \caption{7-scenes data-set volume summary}
 \label{table:7scene_data-set_summery}
\end{table*} 
}
\newcommand{\insertTableCambridgeDatasetSummary}{
\begin{table*}[hbt!]
\centering
 \begin{tabular}{|| l||c c c||} 
  \hline\hline
    Scene & Train set & Test set & Area $[m^2]$ \\
  \hline\hline 
  Great Court       & 1534 & 763 & 8000\\
  King's College   & 1223 & 354 & 5600\\
  Old Hospital     & 898 & 185 & 2000\\
  Shop Facade      & 234 & 106 & 875\\
  St Mary's Church & 1490 & 533 & 4800\\
  Streets          & 3018 & 2926 & 50000\\
 \hline\hline
 \end{tabular}
 \caption{Cambridge Landmark data-set area summary}
 \label{table:cambridge_data-set_summery}
\end{table*}
}
\newcommand{\insertFigColorJitter}{
\begin{figure*}[hbt!]
    \begin{center}
        \begin{subfigure}{0.4\textwidth}
            \includegraphics[width=0.85\linewidth]{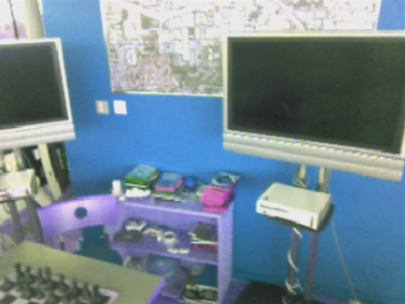}
            \caption{}
        \end{subfigure}
        \begin{subfigure}{0.4\textwidth}
            \includegraphics[width=0.9\linewidth]{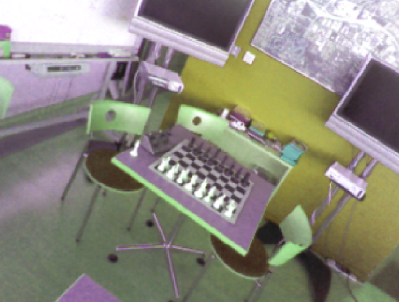}
            \caption{}
        \end{subfigure}
        \begin{subfigure}{0.4\textwidth}
            \includegraphics[width=0.85\linewidth]{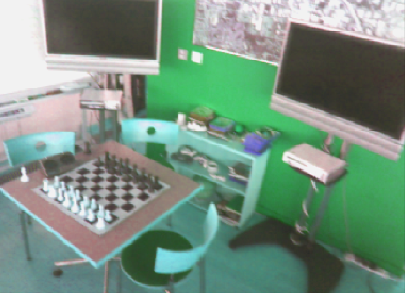}
            \caption{}
        \end{subfigure}
        \begin{subfigure}{0.4\textwidth}
            \includegraphics[width=0.9\linewidth]{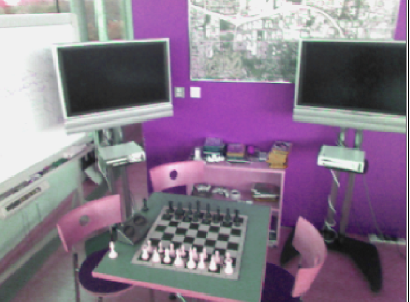}
            \caption{}
        \end{subfigure}
    \end{center}
    \captionsetup{width=.95\textwidth}
    \caption{The figure presents images after color jitter manipulation. All four images picked from the same scene. However, they visually different in lightning and color aspects, making it hard to localize based on visual cues.}
    \label{fig:color-jitter}
\end{figure*}
}
\newcommand{\insertTableDataOverlapRange}{
\begin{table*}[hbt!]
\centering
 \begin{tabular}{|| l||c ||} 
 \hline\hline
  \multicolumn{2}{||c||}{Chess Train Data} \\
  \hline
    Overlap Range &  Number of Pairs \\
  \hline\hline 
  $0^+ - 0.1$ & 228698  \\
  $0.1 - 0.2$ & 148697 \\
  $0.2 - 0.3$ & 128220 \\
  $0.3 - 0.4$ & 119089 \\
  $0.4 - 0.5$ & 107924 \\
  $0.5 - 0.6$ & 97086 \\
  $0.6 - 0.7$ & 87336 \\
  $0.7 - 0.8$ & 105229 \\
  $0.8 - 0.9$ & 11625 \\
 \hline\hline
 \end{tabular}
 \captionsetup{width=.95\textwidth}
 \caption{Example of pre-processing results on chess train data. Table present pairs distribution over overlap range. Any training on the data requires a selection of threshold, which filter pairs lower overlap score. Effectively, different overlap methods lead to major changes on datasets, even when using the same split to train and test.}
 \label{table:overlap_range}
\end{table*}
}
\newcommand{\insertFigArchitecture}{
\begin{figure*}[hbt!]
    \begin{center}
        \begin{subfigure}{0.99\textwidth}
        \begin{center}
           \includegraphics[width=0.8\linewidth]{images/training_pipeline_2.png}
           \caption{Training}
           \label{fig:architecture:training}
        \end{center}
        \end{subfigure}
        \begin{subfigure}{0.9\textwidth}
           \begin{center}
           \includegraphics[width=0.8\linewidth]{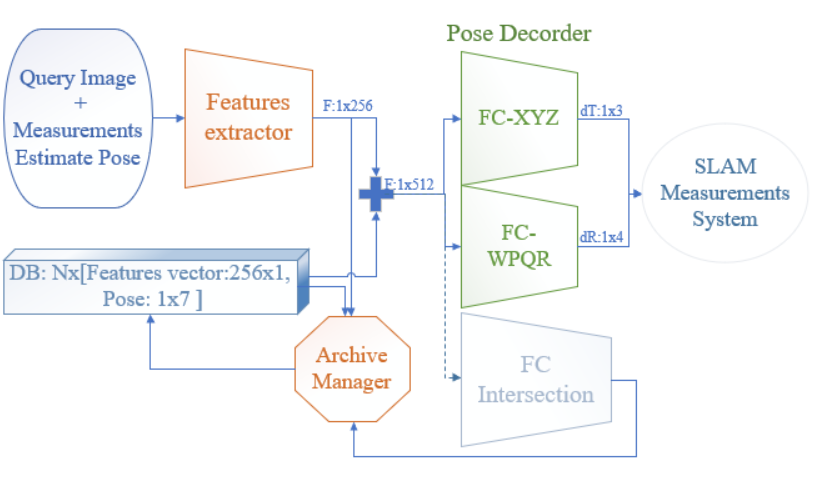}
           \caption{Inference}
           \label{fig:architecture:system}
           \end{center}
        \end{subfigure}
    \end{center}
    \captionsetup{width=.95\textwidth}
    \caption{Diagram of the proposed architecture:\newline 
    {\bf Training}: Two images enter the siamese networks of twin feature extractors and encode into two 256 size feature vectors.  Vectors are concatenated into pose decoder sub-networks to obtain rotation (WPRQ) and translation (XYZ). \newline 
    {\bf Inference}: Given query image the  features extractor produce feature vector. The archive manager compares the feature vector to the database, descriptors are ranked according to their similarity score, and the top-ranked is retrieved.  The pose decoder estimates the relative-pose between the query and the retrieved image. For the new position, we compose the relative-pose on the retrieved image pose.
    }
    \label{fig:architecture}
\end{figure*}
}
\newcommand{\insertTableBeatThemAllRPR}{
\begin{table}[hbt!]
\label{table:relative_7_scene1}
\resizebox{\textwidth}{!}{%
    \begin{tabular}{|l|c|c|c|c||c|} 
    \hline
    & RelocNet \cite{Balntas2018RelocNetCM} & NNnet\cite{Laskar2017CameraRB} & Anchornet \cite{Saha2018ImprovedVR} & CamNet\cite{Ding_2019_ICCV} & Ours \\
    \hline
    Scene & \cellcolor[HTML]{9AFF99}RPR & \cellcolor[HTML]{9AFF99}RPR & \cellcolor[HTML]{9AFF99}RPR & \cellcolor[HTML]{9AFF99}RPR & \cellcolor[HTML]{9AFF99}RPR \\
    \hline
     Chess    & 0.12m, 4.14\dgr & 0.13m,  6.46\dgr & 0.08m, 4.12\dgr & 0.04m, 1.73\dgr & 0.03m, 1.36\dgr\\ \hline
     Fire     & 0.26m, 10.4\dgr & 0.26m, 12.72\dgr & 0.16m, 11.1\dgr & 0.03m, 1.74\dgr & 0.03m, 1.92\dgr\\ \hline
     Heads    & 0.14m, 10.5\dgr & 0.14m, 12.34\dgr & 0.09m, 11.2\dgr & 0.05m, 1.98\dgr & 0.03m, 3.18\dgr\\ \hline
     Office   & 0.18m, 5.32\dgr & 0.21m,  7.35\dgr & 0.11m, 5.38\dgr & 0.04m, 1.62\dgr & 0.03m, 1.75\dgr\\ \hline
     Pumpkin  & 0.26m, 4.17\dgr & 0.24m,  6.35\dgr & 0.14m, 3.55\dgr & 0.04m, 1.64\dgr & 0.03m, 0.98\dgr\\ \hline
     Kitchen  & 0.23m, 5.08\dgr & 0.24m,  8.03\dgr & 0.13m, 5.29\dgr & 0.04m, 1.63\dgr & 0.03m, 1.11\dgr\\ \hline
     Stairs   & 0.28m, 7.53\dgr & 0.27m, 11.80\dgr & 0.21m, 11.9\dgr & 0.04m, 1.51\dgr & 0.03m, 1.27\dgr\\ \hline
     \end{tabular}
 }
 \captionsetup{width=.95\textwidth}
 \caption{The table demonstrates the inadequacy of the existing criterion. We adjusted the overlap parameter to get results that seem better than those of the other methods. However, we emphasize that this is not sufficient to determine which method is preferable.}
 \label{table:beat_them_all:RPR}
\end{table}
}
\newcommand{\insertTableBeatThemAllAPR}{
\begin{table}[hbt!]
\resizebox{\textwidth}{!}{%
\begin{tabular}{|l|l|l|l|l|l|l||l|}
\hline
            & Active search & DSAC++    & Posenet   & \begin{tabular}[c]{@{}l@{}}Posenet\\ Geometric\end{tabular} & MapNet    & CamNet    & Ours  \\ \hline
            & \cellcolor[HTML]{FFFFC7}3D & \cellcolor[HTML]{FFFFC7}3D & \cellcolor[HTML]{FFCE93}APR & \cellcolor[HTML]{FFCE93}APR & \cellcolor[HTML]{FFCE93}APR & \cellcolor[HTML]{9AFF99}RPR & \cellcolor[HTML]{9AFF99}RPR \\ \hline
Chess    & 0.04m,1.96\dgr & 0.02m, 0.5\dgr & 0.32m, 6.60\dgr & 0.13m, 4.48\dgr & 0.08m,  3.25\dgr & 0.04m, 1.73\dgr & 0.03m, 1.36\dgr\\ \hline
Fire     & 0.03m,1.53\dgr & 0.02m, 0.9\dgr & 0.47m, 14.0\dgr & 0.27m, 11.3\dgr & 0.27m, 11.69\dgr & 0.03m, 1.74\dgr & 0.03m, 1.92\dgr\\ \hline
Heads    & 0.02m,1.45\dgr & 0.01m, 0.8\dgr & 0.30m, 12.2\dgr & 0.17m, 13.0\dgr & 0.18m, 13.25\dgr & 0.05m, 1.98\dgr & 0.03m, 3.18\dgr\\ \hline
Office   & 0.09m,3.61\dgr & 0.03m, 0.7\dgr & 0.48m, 7.24\dgr & 0.19m, 5.55\dgr & 0.17m,  5.15\dgr & 0.04m, 1.62\dgr & 0.03m, 1.75\dgr\\ \hline
Pumpkin  & 0.08m,3.10\dgr & 0.04m, 1.1\dgr & 0.49m, 8.12\dgr & 0.26m, 4.75\dgr & 0.22m,  4.02\dgr & 0.04m, 1.64\dgr & 0.03m, 0.98\dgr\\ \hline
Kitchen  & 0.07m,3.37\dgr & 0.04m, 1.1\dgr & 0.58m, 8.34\dgr & 0.23m, 5.35\dgr & 0.23m,  4.93\dgr & 0.04m, 1.63\dgr & 0.03m, 1.11\dgr\\ \hline
Stairs   & 0.03m,2.22\dgr & 0.09m, 2.6\dgr & 0.48m, 13.1\dgr & 0.35m, 12.4\dgr & 0.30m, 12.08\dgr & 0.04m, 1.51\dgr & 0.03m, 1.27\dgr\\ \hline
\end{tabular}}
\captionsetup{width=.95\textwidth}
\caption{Misleading comparison to other methods (APR: absolute-pose-regression, RPR: relative-pose-regression, 3D-classical) using the standard median metric on 7-scenes data-set.}
\label{table:beat_them_all}
\end{table}
}
\newcommand{\insertFigNaivevsGood}{
\begin{figure}[hbt!]
    \begin{center}
        \begin{subfigure}{0.5\textwidth}
            \includegraphics[width=0.9\linewidth]{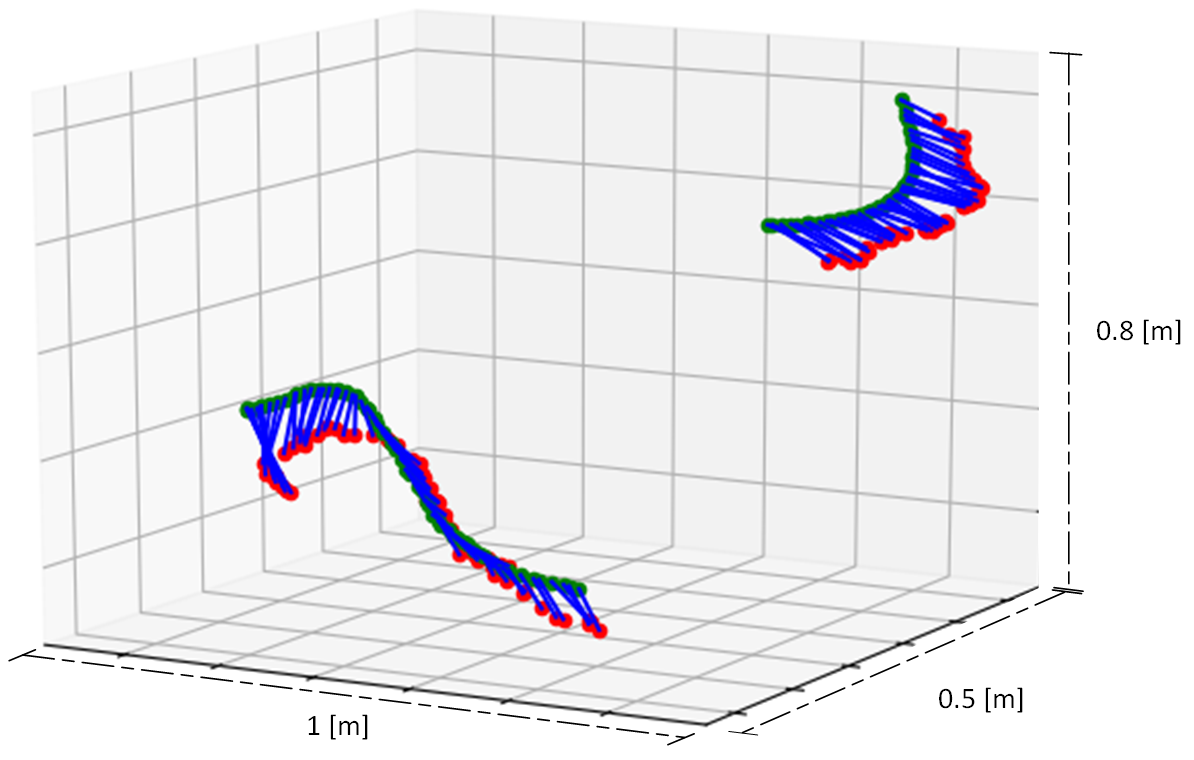}
            \caption{Non-naive model. Mean translation error 0.11}
        \end{subfigure}
        \begin{subfigure}{0.47\textwidth}
            \includegraphics[width=0.9\linewidth]{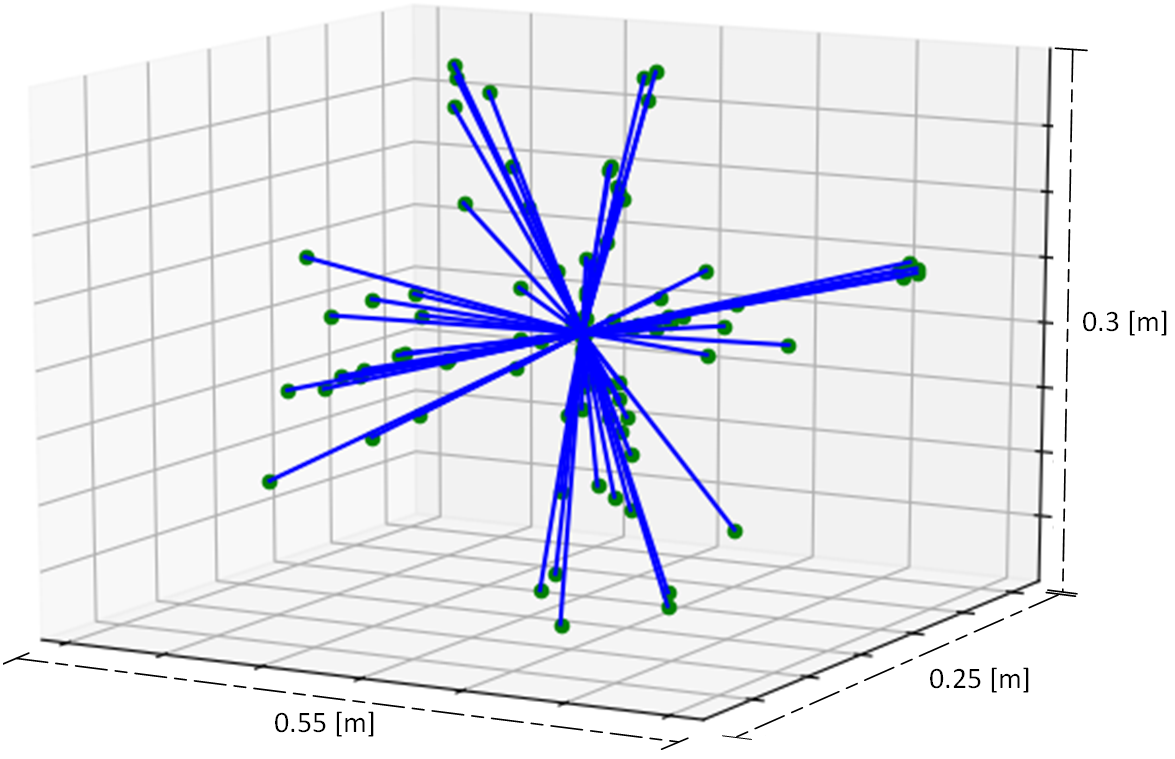}
                \caption{Naive model. Mean translation error 0.10 }
        \end{subfigure}
    \end{center}
    \captionsetup{width=.95\textwidth}
    \caption{Consider only the numerical results the naive model (b) is better than the non-naive one (a), after examination of the visual results, it seems clearly that model (b) is worthless, when model (a) learned to predict the pose from the training data. The mistake stems from not taking the subspace volume into account. Not that in this illustration only translation error have been discussed. The same argument holds for rotation error.}
    \label{fig:Naive_vs_good}
\end{figure}
}

\newcommand{\tablescenesdiameter}{
\begin{center}
\resizebox{0.9\textwidth}{!}{\begin{tabular}{|l|p{1cm}|p{1cm}|p{1cm}|p{1cm}|p{1cm}|}
    \hline
    \multirow{2}{*}{} & \multicolumn{5}{c|}{Overlap Threshold} \\ \cline{2-6} 
     &     0.2  & 0.4  & 0.6  & 0.8  & 0.9 \\
    \hline
    Chess   & 1.95 & 1.91 & 1.54 & 0.71 & 0.57 \\
    Fire    & 1.65 & 1.16 & 0.80 & 0.77 & 0.68 \\
    Heads   & 2.13 & 1.93 & 1.13 & 0.57 & 0.45 \\
    Office  & 2.79 & 2.65 & 2.28 & 0.92 & 0.42 \\
    Pumpkin & 1.24 & 1.34 & 1.42 & 0.80 & 0.26 \\
    Kitchen & 3.49 & 2.47 & 1.76 & 0.88 & 0.38 \\
    Stairs  & 0.91 & 0.75 & 0.60 & 0.50 & 0.37 \\
    \hline
\end{tabular}}
\end{center}}

\newcommand{\insertTableOverlapDiameter}{
\begin{table}[hbt!]
\begin{center}
    \begin{subfigure}[t]{0.8\linewidth}
        \captionsetup{width=.95\textwidth}
        \tablescenesdiameter
        \caption{Demonstrated on 7-scenes \cite{Shotton2013SceneCR} data-set.}
        \label{table:overlap_define_subspace_diameter}
        \label{table:overlap_diameter:7-scenes}
        \end{subfigure} 
    \begin{subfigure}[t]{0.8\linewidth}
        \captionsetup{width=.95\textwidth}
        \tablecambridgediameter
        \caption{Demonstrated on Cambridge \cite{PoseNet2015} data-set.}
        \label{table:overlap_diameter:Cambridge}
    \end{subfigure} 
    \end{center}
    \captionsetup{width=.95\textwidth}
    \caption{Relative poses regression subspace diameter in meters as a function of overlap. The subspace is created by the span of all relative-poses with enough overlap. The diameters are correlated to the overlap threshold, while the scene volume has little effect.}
    \label{table:max_overlap_allowed}
    \label{table:overlap_diameter}
\end{table}
}
\newcommand{\insertTableOverlapDiameterSmall}{
\begin{table}
\captionsetup{width=.95\textwidth}
\begin{center}
    \begin{subtable}[t]{0.42\linewidth}
        \captionsetup{width=.95\textwidth}
        \tablescenesdiameter
        \caption{Demonstrated on 7-scenes\cite{Shotton2013SceneCR} data-set.}
        \label{table:overlap_define_subspace_diameter_small}
        \label{table:overlap_diameter:7-scenes_small}
        \end{subtable} 
    \begin{subtable}[t]{0.48\linewidth}
        \captionsetup{width=.95\textwidth}
        \tablecambridgediameter
        \caption{Demonstrated on Cambridge\cite{PoseNet2015} data-set.}
         \label{table:overlap_diameter:Cambridge_small}
    \end{subtable} 
    \end{center}
    \captionsetup{width=.95\textwidth}
    \caption{Relative poses regression subspace diameter in meters as a function of overlap. The subspace created by the span of all relative-poses with enough overlap. The diameters are correlated to overlap threshold, while the scene volume almost not matter.}
    \label{table:overlap_diameter_small}
\end{table}
}
\newcommand{\insertFigAUC}{
\begin{figure}[hbt!]
\begin{center}
    \begin{subfigure}[t]{0.38\textwidth}
        \includegraphics[width=0.99\linewidth]{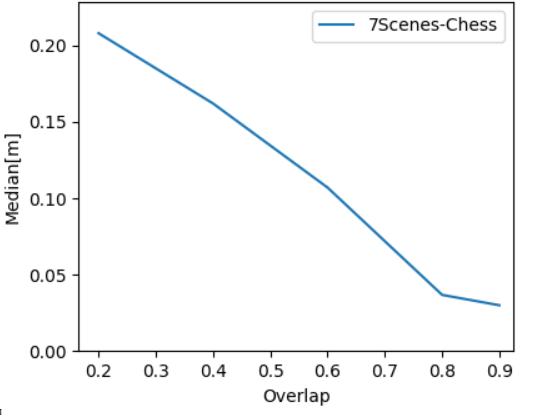}
        \caption{ }
        \label{fig:diameter_accuracy_corr}
        \end{subfigure} 
    \begin{subfigure}[t]{0.48\textwidth}
        \includegraphics[width=0.99\linewidth]{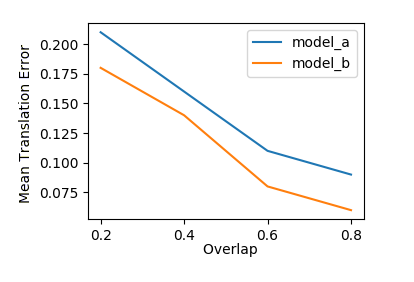}
        \caption{}
        \label{fig:AUC_single}
    \end{subfigure} 
    \captionsetup{width=.95\textwidth}
    \caption{
    (a) Accuracy across different overlaps on 7-scenes data-set. We can see clearly the correlation, when overlap increase accuracy decrease. (b)  Two trained models tested on the same data. We can see that Model b performs better than Model a since it valid for any overlap score. To evaluate and compare in cases there is no single model that is always better, we turned to the area under the curve (AUC) approach.}
\end{center}
\end{figure}
}
\newcommand{\insertFigGraphTwoModeld}{
\begin{figure}[hbt!]
    \begin{center}
    \includegraphics[width=0.6\linewidth]{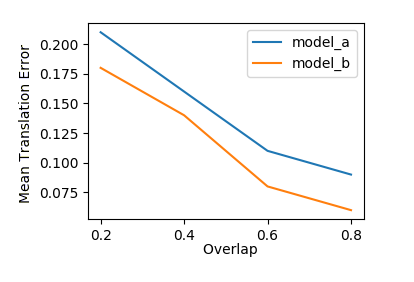}
    \end{center}
    \captionsetup{width=.95\textwidth}
    \caption{ Two trained models tested on the same data. We can say model b performs better than model a since it valid for any overlap score. To measure how much better or to compare in cases there is no single model that is always better, we justified comparison on area under the curve (AUC).}
    \label{fig:AUC}
    \label{fig:graph_2_modeld}
\end{figure}
}
\newcommand{\insertFigMedianAcrossOverlaps}{
\begin{figure}[hbt!]
\begin{center}
        \includegraphics[width=0.80\linewidth]{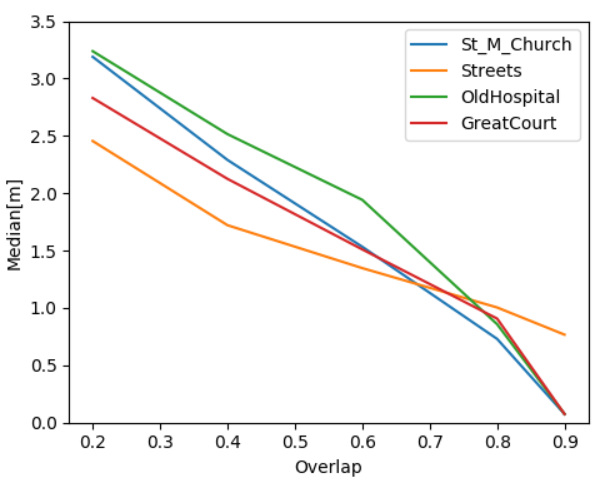}
        \captionsetup{width=.95\textwidth}
        \caption{Median accuracy across different overlaps on Cambridge data-set.}
        \label{fig:MedianAcrossOverlaps}
\end{center}
\end{figure}

}
\newcommand{\insertFigMAPEAcrossOverlaps}{
\begin{figure}[hbt!]
\begin{center}
    \begin{subfigure}[t]{0.38\textwidth}
        \includegraphics[width=0.99\linewidth]{images/graph_same_model_Cam_median.png}
        \caption{Median accuracy across different overlaps on Cambridge data-set.}
        \end{subfigure} 
    \begin{subfigure}[t]{0.38\textwidth}
        \includegraphics[width=0.99\linewidth]{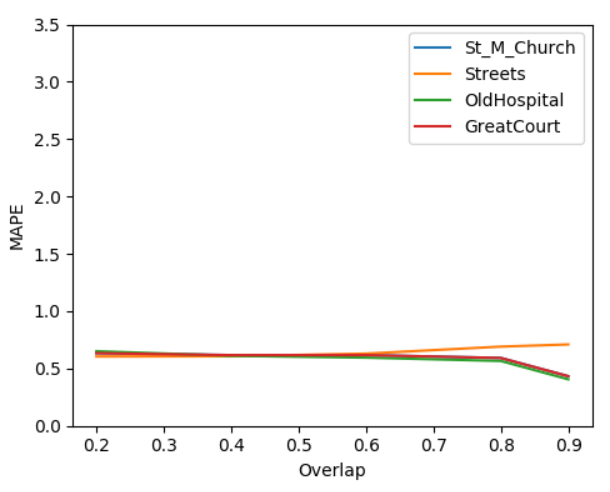}
        \caption{MAPE across different overlaps on Cambridge data-set.}
    \end{subfigure} 
    \end{center}
    \captionsetup{width=.95\textwidth}
    \caption{Translation error using median (a) and  using MAPE (b) across different overlaps.
    } 
\end{figure}
}
\newcommand{\insertFigNaiveEstimationRMSE}{
\begin{figure}[hbt!]
    \begin{center}
        \begin{subfigure}{0.5\textwidth}
            \includegraphics[width=0.9\linewidth]{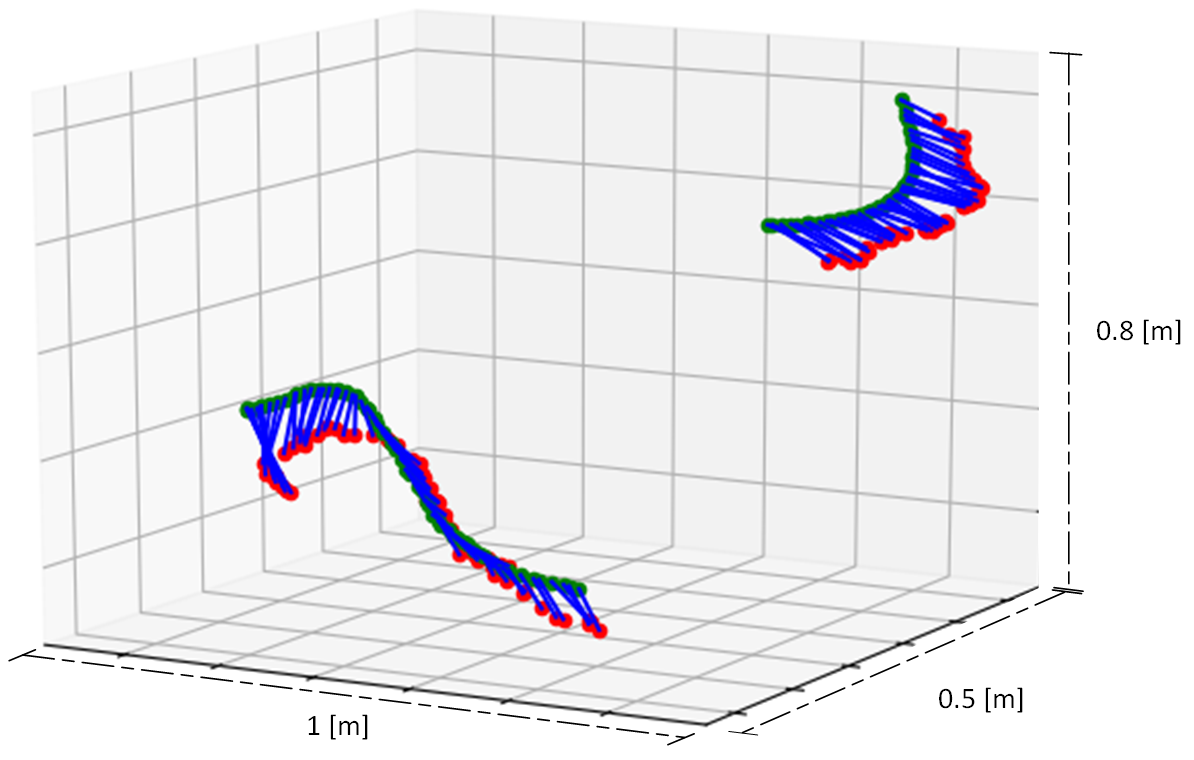}
            \caption{\newline
                \begin{tabular}{|l|}
                    \hline 
                    {\bf model a} \\
                    \hline 
                    15 epochs of training\\
                    100 pairs of images\\
                    Minimum overlap of 0.4 \\   
                    Subspace size: $1 \times 0.5 \times 0.8= 0.4[m^3$] \\
                    \hline
                    Mean translation : 0.11 [m]       \\
                    Mean Rotation: 3.51 [degree]   \\
                    \hline
                \end{tabular}
                }
        \end{subfigure}
        \begin{subfigure}{0.47\textwidth}
            \includegraphics[width=0.9\linewidth]{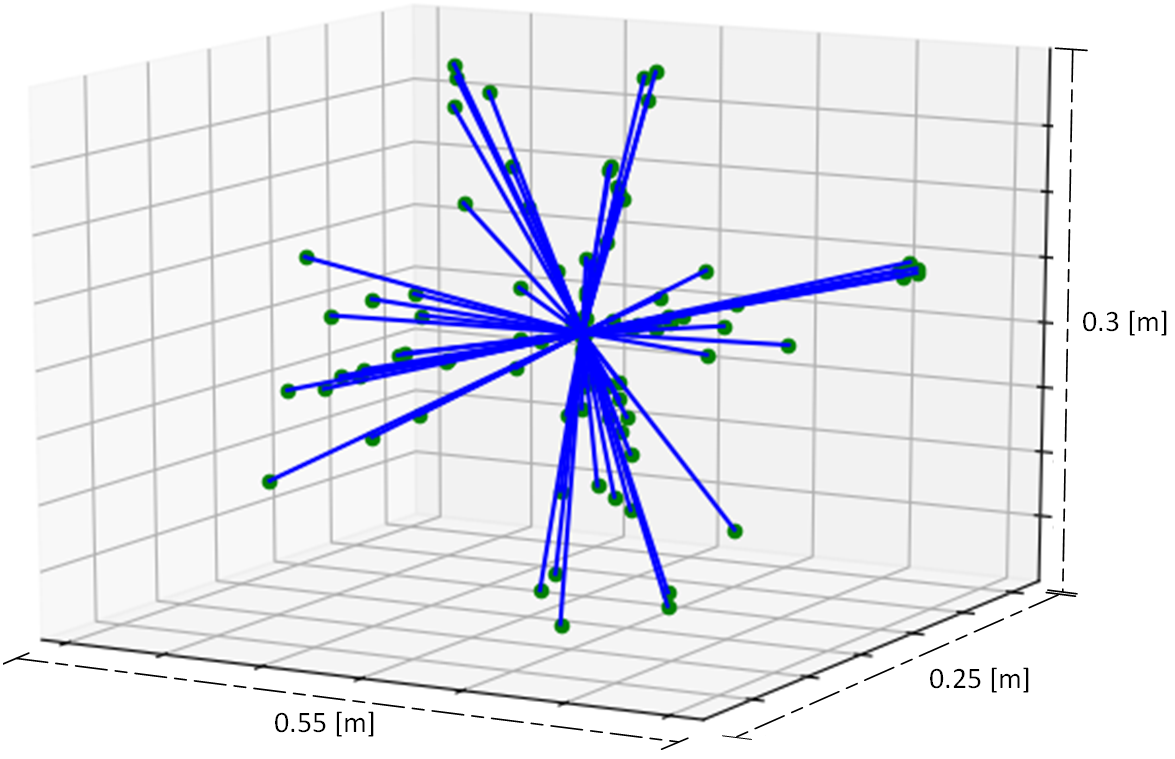}
                \caption{\newline
                \begin{tabular}{|l|}
                    \hline 
                    {\bf model b} \\
                    \hline 
                    1 epoch of training\\
                    100 pairs of images\\
                    Minimum overlap of 0.95 \\  
                    Subspace sizes: $0.6 \times 0.25 \times 0.3= 0.045[m^3$] \\
                    \hline
                    Mean translation : 0.10 [m]       \\
                    Mean Rotation: 3.39 [degree]   \\
                    \hline
                \end{tabular}
                }
        \end{subfigure}
    \end{center}
    \captionsetup{width=.95\textwidth}
    \caption{The figures present the translations error of two models. Red and green points are ground truths and estimates, respectively. Considering only the numerical results the naive model (b) is better. However, after visual examination, it is clear that model (b) is worthless, while model (a) learned to predict the pose. The mistake stems from not taking the subspace volume into account.}
        
    \label{fig:Naive-Estimation-RMSE}
\end{figure}
}
\newcommand{\tablecambridgediameter}{
\begin{center}
\resizebox{0.95\textwidth}{!}{\begin{tabular}{|l|p{1cm}|p{1cm}|p{1cm}|p{1cm}|p{1cm}|}
    \hline
    \multirow{2}{*}{} & \multicolumn{5}{c|}{Overlap Threshold} \\ \cline{2-6} 
     &     0.2  & 0.4  & 0.6  & 0.8  & 0.9 \\
    \hline
    Kings Collage   & 6.82 & 4.87 & 3.51 & 2.50 & 1.26 \\
    Great Court     & 6.54 & 5.00 & 3.59 & 2.39 & 1.58 \\
    Old Hospital    & 8.08 & 6.71 & 5.16 & 2.73 & 0.89 \\
    St M. Church    & 8.43 & 7.21 & 5.61 & 2.40 & 0.69 \\
    Streets         & 6.48 & 4.78 & 3.56 & 2.45 & 1.14 \\
    Shop Facade      & 6.08 & 4.28 & 3.79 & 2.42 & 1.43 \\
    \hline

\end{tabular}}
\end{center}}
\newcommand{\insertFigVisualizeMetricies}{
\begin{figure}[t]
    \begin{subfigure}{0.99\textwidth}
        \begin{center}
        \includegraphics[width=0.50\linewidth]{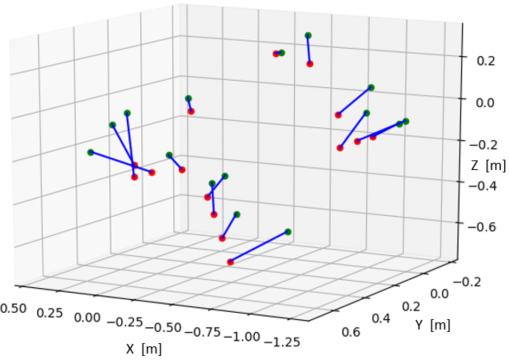}
        \captionsetup{width=0.8\textwidth}
        \caption{Translations error - Each point represent relative-pose translation in 3D terms. Green and red points are ground truth and estimation respectively. Blue lines represent the errors in terms of euclidean distance.}
        \label{fig:visualize_metric:median}
        \end{center}
    \end{subfigure} 
    \begin{subfigure}{0.99\textwidth}
        \begin{center}
        \includegraphics[width=0.50\linewidth]{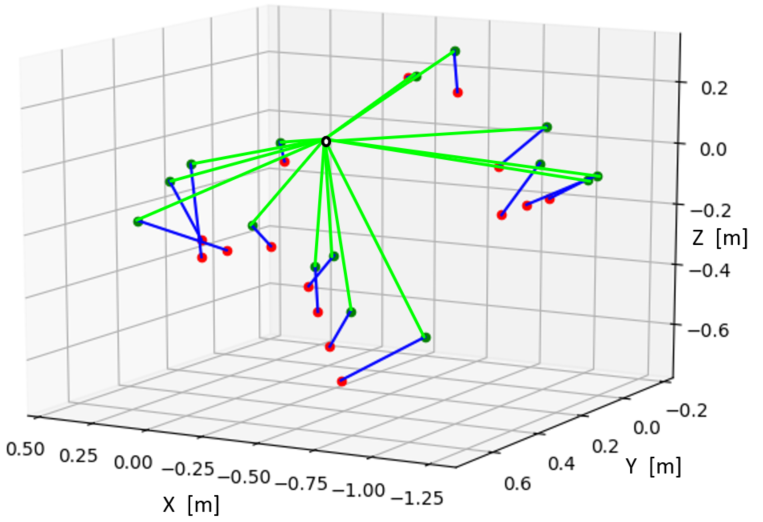}
        \captionsetup{width=0.8\textwidth}
        \caption{Mean absolute percentage error (MAPE). The additional green lines are the normalization factor used. MAPE normalization factor is norm of ground trues point.}
        \label{fig:visualize_metric:MAPE}
        \end{center}
    \end{subfigure} 
    \begin{subfigure}{0.99\textwidth}
        \begin{center}
        \includegraphics[width=0.5\linewidth]{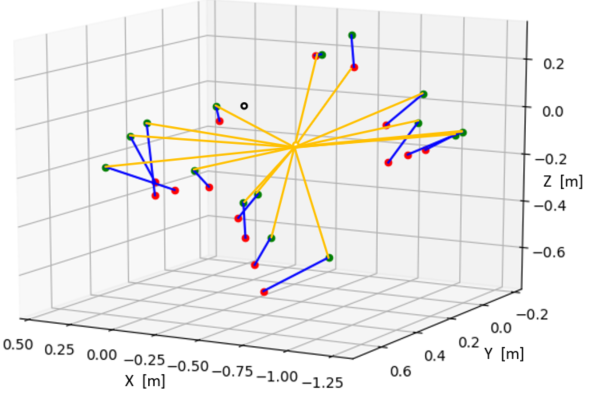}
        \captionsetup{width=0.8\textwidth}
        \caption{Mean Absolute Scale Error (MASE). Normalization factor in yellow is the naive model error.}
        \label{fig:visualize_metric:MASE}
        \end{center}
    \end{subfigure} 
    \captionsetup{width=.95\textwidth}
    \caption{Figure demonstrate the normalization factors of MAPE and MASE metrics on the translations errors.} 
    \label{fig:visualize_metric} 
    
\end{figure}
}

\newcommand{\insertFigAgnosicVloumeMethodsIII}{
\begin{figure}[hbt!]
    \begin{center}
        \includegraphics[width=0.99\linewidth]{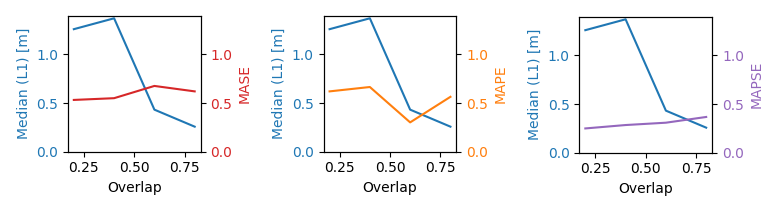}
    \end{center}
    \captionsetup{width=.95\textwidth}
    \caption{An illustration of using volume-agnostic metrics vs median metric under L1 norm. The blue line at each plot is the median, when the other line in each plot is one of the 3 proposed metrics. It can be clearly seen that the proposed method is robust to different overlaps. 
}
    \label{fig:agnosic_vloume_methods_3}
\end{figure}
}
\newcommand{\insertFigAgnosicVloumeMethodsIV}{
\begin{figure}[hbt!]
    \begin{center}
        \includegraphics[width=0.99\linewidth]{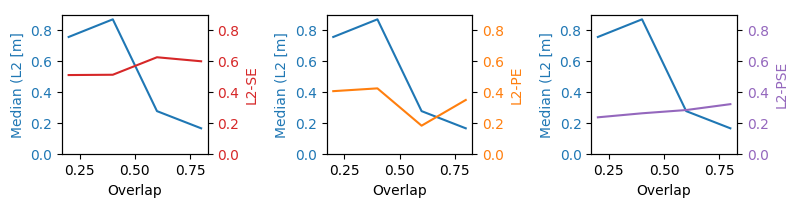}
    \end{center}
    \captionsetup{width=.95\textwidth}
    \caption{Proposed metrics are agnostic to norm selection. Unlike Fig.~\ref{fig:agnosic_vloume_methods_3} here we use $L2$. It can be seen that the proposed metrics are robust to \textbf{norm selection}.} 
    \label{fig:agnosic_vloume_methods_4}
\end{figure}
}
\newcommand{\insertTableCompareSubspaceDiameterSmall}{
    \begin{table}[hbt!]
    \begin{center}
        \begin{tabular}{|l|c|c|c|c|}
            \hline
                      &\multicolumn{2}{c|}{AnchorNet}   & \multicolumn{2}{c|}{Ours best}    \\
            \hline
                  &Accuracy    &\begin{tabular}[c]{@{}l@{}}Subspace\\  Diameter\end{tabular} 
                  &Accuracy    &\begin{tabular}[c]{@{}l@{}}Subspace\\  Diameter\end{tabular} \\
            \hline 
            Chess       &0.08m, 4.12\dgr  &0.19  &0.07m, 1.63\dgr  &0.19 \\
            Fire        &0.16m, 11.1\dgr  &0.14  &0.02m, 1.45\dgr  &0.14 \\
            Heads       &0.09m, 11.2\dgr  &0.05  &0.02m, 1.73\dgr  &0.05 \\
            Office      &0.11m, 5.38\dgr  &0.18  &0.02m, 1.53\dgr  &0.18 \\
            Pumpkin     &0.14m, 3.55\dgr  &0.11  &0.02m, 1.09\dgr  &0.11 \\
            Red Kitchen &0.13m, 5.29\dgr  &0.22  &0.02m, 1.14\dgr  &0.22 \\
            Stairs      &0.21m, 11.9\dgr  &0.17  &0.01m, 0.74\dgr  &0.17 \\
            \hline 
            \end{tabular}
        \captionsetup{width=.95\textwidth}
        \caption{ Comparison to \cite{Saha2018ImprovedVR} with same subspace diameter using same overlap method }
        \label{tab:compare_sub_space_diameter}
    \end{center}
    \end{table}
}
\newcommand{\insertTableCompareSubspaceDiameter}{

    \begin{table}[h]
        \begin{center}
        \begin{tabular}{|l|c|c|c|c|c|c|}
            \hline
                      &\multicolumn{2}{c|}{AnchorNet}   & \multicolumn{2}{c|}{Ours best}   & \multicolumn{2}{c|}{Ours chess model}      \\
            \hline
                  &Accuracy    &\begin{tabular}[c]{@{}l@{}}Subspace\\  Diameter\end{tabular} 
                  &Accuracy    &\begin{tabular}[c]{@{}l@{}}Subspace\\  Diameter\end{tabular} 
                  &Accuracy    &\begin{tabular}[c]{@{}l@{}}Subspace\\  Diameter\end{tabular} \\
            \hline \hline
            chess       &0.08m, 4.12\dgr  &0.19  &0.07m, 1.63\dgr  &0.19  & 0.07m, 1.63\dgr    & 0.19 \\
            fire        &0.16m, 11.1\dgr  &0.14  &0.02m, 1.45\dgr  &0.14  & 0.16m, 6.25\dgr    & 0.14 \\
            Heads       &0.09m, 11.2\dgr  &0.05  &0.02m, 1.73\dgr  &0.05  & 0.07m, 3.43\dgr    & 0.05 \\
            Office      &0.11m, 5.38\dgr  &0.18  &0.02m, 1.53\dgr  &0.18  & 0.12m, 4.35\dgr    & 0.18 \\
            Pumpkin     &0.14m, 3.55\dgr  &0.11  &0.02m, 1.09\dgr  &0.11  & 0.09m, 3.96\dgr    & 0.11 \\
            Red Kitchen    &0.13m, 5.29\dgr  &0.22  &0.02m, 1.14\dgr  &0.22  & 0.10m, 4.06\dgr    & 0.22 \\
            Stairs      &0.21m, 11.9\dgr  &0.17  &0.01m, 0.74\dgr  &0.17  & 0.10m, 2.54\dgr    & 0.17 \\
            \hline 
        \end{tabular}
        \captionsetup{width=.95\textwidth}
        \caption{Comparison to \cite{Saha2018ImprovedVR} given the subspace diameter. Note that in the third double-column we train only on one scene and yet are able to generalize to other scenes as well.}
        \label{tab:compare_sub_space_diameter_plus_generalzation}
    \end{center}
    \end{table}
}
\newcommand{\tablemeidianlay}{
\begin{center}
\begin{tabular}{|l|c|c|c|c|}
\hline
Criterion  & Translation & Rotation & Translation & Rotation \\
\hline
Mean   & 0.16 [m] & 3.43\dgr & 0.21 [m] & 5.01 \\
Median & 0.13 [m] & 2.78\dgr & 0.12 [m]& 2.49 \\
APE L1 & 0.23     &      & 0.31 &      \\
MAPE   & 0.24     &      & 0.31 &      \\
MASE   & 0.12     &      & 0.15 &      \\
MAPSE  & 0.17     &      & 0.23 &      \\   
\hline
\end{tabular}
\end{center}
}
\newcommand{\insertTableGlobalExperiments}{
\begin{table*}[hbt!]
\begin{tabular}{||>{\columncolor[gray]{0.8}}l||p{1.52cm}|p{1.52cm}|p{1.52cm}|p{1.52cm}|p{1.52cm}|p{1.52cm}|p{1.52cm}||} 
\hline\hline
 \rowcolor[gray]{0.8} & Chess & Stairs & kitchen & Heads & Fire & Office & Pumpkin \\ \hline
Chess   &\cellcolor[gray]{0.9} min 0.99, max 1.00 & & & & & & \\  \hline
Stairs  &min 0.72, max 0.74 &\cellcolor[gray]{0.9}min 0.99, max 1.00 & & & & & \\ \hline  
kitchen &min 0.54, max 0.58 &min 0.69, max 0.78 &\cellcolor[gray]{0.9}min 0.99, max 1.00 & & & & \\  \hline
Heads   &min 0.18, max 0.01 &min 0.34, max 0.48 &min 0.57, max 0.69 &\cellcolor[gray]{0.9}min 0.97, max 1.00  & & & \\   \hline
Fire    &min 0.60, max 0.80 &min 0.88, max 0.90 &min 0.81, max 0.91 &min 0.51, max 0.72  &\cellcolor[gray]{0.9}min 0.96, max 1.00  & & \\  \hline
Office  &min 0.78, max 0.91 &min 0.65, max 0.82 &min 0.78, max 0.86 &min 0.24, max 0.39  &min 0.69, max 0.86  &\cellcolor[gray]{0.9}min 0.95, max 1.00  & \\  \hline
Pumpkin &min 0.37, max 0.46 &min 0.76, max 0.79 &min 0.83, max 0.89 &min 0.77, max 0.88  &min 0.81, max 0.94  &min 0.81, max 0.94  &\cellcolor[gray]{0.9}min 0.99, max 1.00  \\   
\hline\hline 
 \end{tabular}
 \captionsetup{width=.95\textwidth}
 \caption{we took sets of 20 images from each scene on 7-scenes data set, run each image trough the features extractor trained only on one scene, and analyze the correlation by dot product.  the statistics of the results represent in the table.}
\label{table:Global_Experiments}
\end{table*}
}
\newcommand{\insertTableGlobalExperimentsLimits}{
\begin{table*}[hbt!]

\begin{tabular}{||>{\columncolor[gray]{0.8}}l||p{1.52cm}|p{1.52cm}|p{1.52cm}|p{1.52cm}|p{1.52cm}|p{1.52cm}|p{1.52cm}||} 
\hline\hline
 \rowcolor[gray]{0.8} & Chess & Stairs & kitchen & Heads & Fire & Office & Pumpkin \\ \hline
Chess   &\cellcolor[gray]{0.9} min 0.95, max 1.00 & & & & & & \\  \hline
Stairs  &min 0.71, max 0.83 &\cellcolor[gray]{0.9}min 0.99, max 1.00 & & & & & \\ \hline  
kitchen &min 0.55, max 0.80 &min 0.69, max 0.92 &\cellcolor[gray]{0.9}min 0.95, max 1.00 & & & & \\  \hline
Heads   &min 0.28, max 0.28 &min 0.21, max 0.55 &min 0.42, max 0.80 &\cellcolor[gray]{0.9}min 0.93, max 1.00  & & & \\   \hline
Fire    &min 0.55, max 0.85 &min 0.75, max 0.94 &min 0.68, max \textcolor{red}{0.97} &min 0.28, max 0.75  &\cellcolor[gray]{0.9}min 0.81, max 1.00  & & \\  \hline
Office  &min 0.50, max \textcolor{red}{0.96} &min 0.59, max 0.88 &min 0.78, max \textcolor{red}{0.98} &min 0.12, max 0.74  &min 0.51, max 0.89  &\cellcolor[gray]{0.9}min 0.78, max 1.00  & \\  \hline
Pumpkin &min 0.38, max 0.73 &min 0.76, max 0.89 &min 0.82, max \textcolor{red}{0.95} &min 0.57, max 0.89  &min 0.70, max \textcolor{red}{0.97}  &min 0.56, max \textcolor{red}{0.85}  &\cellcolor[gray]{0.9}min 0.97, max 1.00  \\   
\hline\hline 
 \end{tabular}
    \captionsetup{width=.95\textwidth}
    \caption{Similar representation as in Table \ref{table:Global_Experiments}, However, here we increase the sampling sets to include hundreds of images. Red highlights are the results that may create localization confusion} 
\label{table:Global_Experiments_Limits}
\end{table*}
}
\newcommand{\insertFigConfutionFirePumpkin}{
\begin{figure}[hbt!]
    \begin{center}
        \includegraphics[width=0.99\linewidth]{images/confution_fire_pumpkin.png}
    \captionsetup{width=.95\textwidth}
    \caption{Images from pumpkin and fire scenes of 7-scenes dataset. model that trained only on chess scene extract close feature vector for them.     } 
    \label{fig:confution_fire_pumpkin}
    \end{center}
\end{figure}
}